\documentclass[letterpaper]{article} % DO NOT CHANGE THIS
\usepackage{aaai23}  % DO NOT CHANGE THIS
\usepackage{times}  % DO NOT CHANGE THIS
\usepackage{helvet}  % DO NOT CHANGE THIS
\usepackage{courier}  % DO NOT CHANGE THIS
\usepackage[hyphens]{url}  % DO NOT CHANGE THIS
\usepackage{graphicx} % DO NOT CHANGE THIS
\usepackage{natbib}  % DO NOT CHANGE THIS AND DO NOT ADD ANY OPTIONS TO IT
\usepackage{caption} % DO NOT CHANGE THIS AND DO NOT ADD ANY OPTIONS TO IT
\usepackage{algorithm}
\usepackage{amsmath}
\usepackage{amsthm}
\usepackage{bbm}
\usepackage{mathtools}
\usepackage{algpseudocode}
\usepackage{verbatim}
\usepackage{todonotes}
\usepackage{subcaption}

\usepackage{float}
\newfloat{algorithm}{t}{lop} %floating the algorithm

\usepackage{newfloat}
\usepackage{listings}
\DeclareCaptionStyle{ruled}{labelfont=normalfont,labelsep=colon,strut=off} % DO NOT CHANGE THIS
\floatstyle{ruled}
\newfloat{listing}{tb}{lst}{}
\floatname{listing}{Listing}

\newcommand{\rollout}{\text{rollout}}
\newcommand{\Pdata}{\mathbb{P}_{\text{data}}}
\newcommand{\ExP}[2]{\Eop_{#1}\left[{#2}\right]}

\newcommand{\cub}{{\texttt{CUB}}}
\newcommand{\chexpert}{{\texttt{CHEXPERT}}}
\newcommand{\oai}{{\texttt {OAI}}}

\newcommand{\greedy}{{\texttt{Greedy}}}
\newcommand{\skyline}{{\texttt{Skyline}}}
\newcommand{\random}{{\texttt{Random}}}
\newcommand{\AFA}{{\texttt{AFA}}}
\newcommand{\ours}{{\texttt{CooP}}}

\newcommand{\pol}{\psi}
\newcommand{\Sc}{\overline{S}}
\newcommand{\cost}{q}
\newcommand{\br}[1]{\left(#1\right)}
\newcommand{\R}{\mathbb{R}}

\DeclareMathOperator*{\Eop}{\mathbb{E}}

\usepackage[utf8]{inputenc} % allow utf-8 input
\usepackage{hyperref}       % hyperlinks
\usepackage{booktabs}       % professional-quality tables
\usepackage{amsfonts}       % blackboard math symbols
\usepackage{nicefrac}       % compact symbols for 1/2, etc.
\usepackage{microtype}      % microtypography
\usepackage{xcolor}         % colors
\newcommand{\ctoy}{$\mathcal{C} \rightarrow \mathcal{Y}$}
\newcommand{\ftoc}{$\mathcal{X} \rightarrow \mathcal{C}$}
\newcommand{\ftoctoy}{$\mathcal{X} \rightarrow \mathcal{C} \rightarrow \mathcal{Y}$}

\DeclareMathOperator*{\argmax}{\text{argmax}}

\title{Interactive Concept Bottleneck Models}

\author{
    Kushal Chauhan,
    Rishabh Tiwari,
    Jan Freyberg,
    Pradeep Shenoy,
    Krishnamurthy Dvijotham\thanks{Corresponding author}
}
\affiliations{
    Google Research\\
    \{kushalchauhan, rishabhtiwari,janfreyberg,shenoypradeep,dvij\}@google.com

}

\begin{document}

\maketitle

\begin{abstract}
Concept bottleneck models (CBMs) are interpretable neural networks that first predict labels for human-interpretable concepts relevant to the prediction task, and then predict the final label based on the concept label predictions.  We extend CBMs to interactive prediction settings where the model can query a human collaborator for the label to some concepts. We develop an interaction policy that, at prediction time, chooses which concepts to request a label for so as to maximally improve the final prediction. We demonstrate that a simple policy combining concept prediction uncertainty and influence of the concept on the final prediction achieves strong performance and outperforms static approaches as well as active feature acquisition methods proposed in the literature. We show that the interactive CBM can achieve accuracy gains of 5-10\% with only 5 interactions over competitive baselines on the Caltech-UCSD Birds, CheXpert and OAI datasets.
\end{abstract}
%Datasets

\section{Introduction}

\begin{figure}[!t]
    \centering
    \begin{subfigure}[b]{\linewidth}
         \centering
         \includegraphics[width=0.9\linewidth]{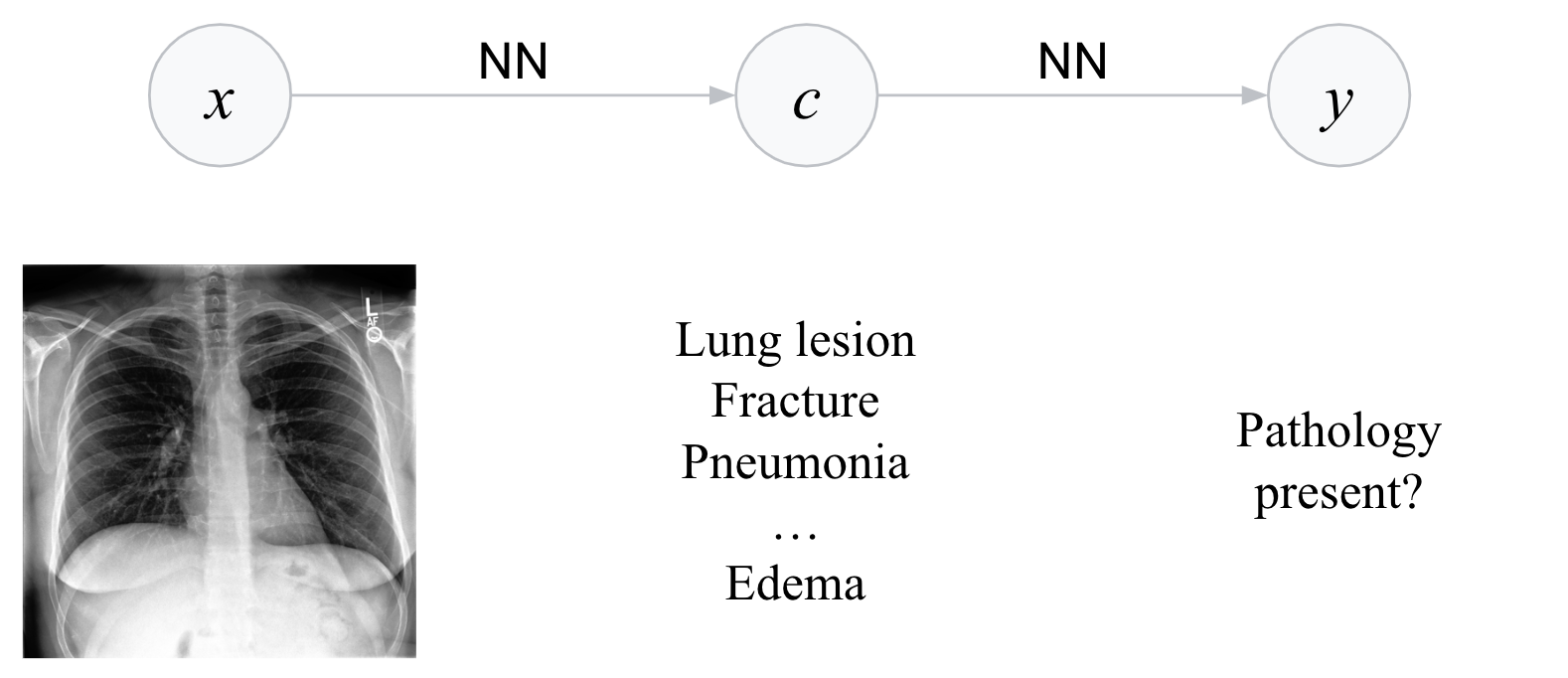}
         \caption{Concept Bottleneck Model}
     \end{subfigure}
    
    \begin{subfigure}[b]{\linewidth}
         \centering
         \includegraphics[width=0.9\linewidth]{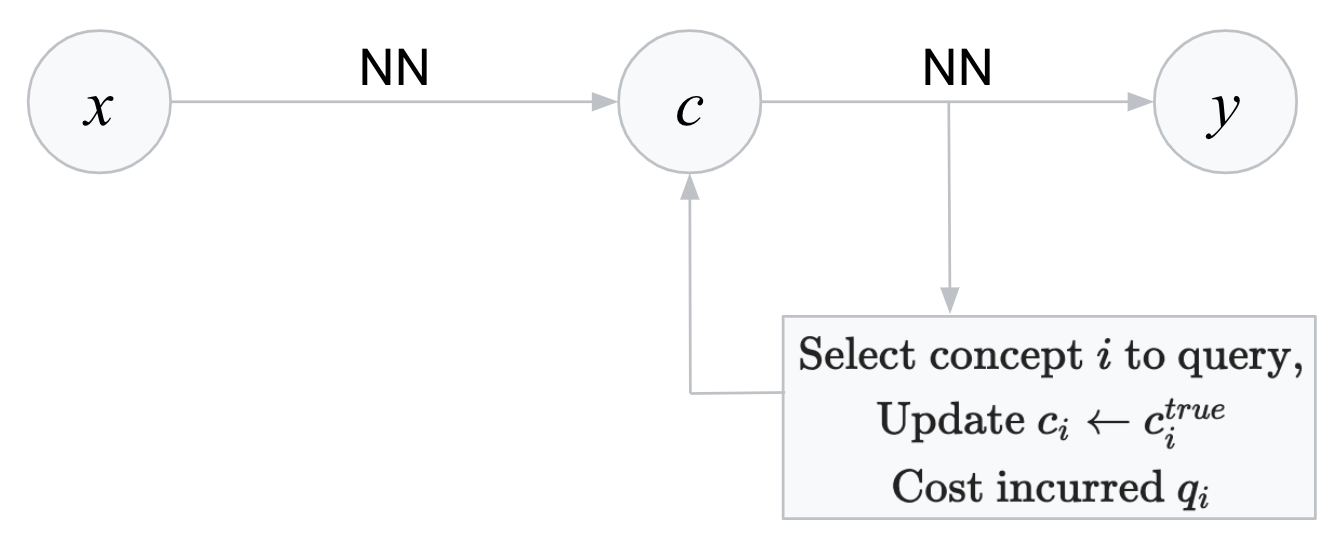}
         \caption{Interactive Prediction}
    \end{subfigure}

    \caption{Interactive Prediction: Panel (a) shows a concept bottleneck model~\cite{pmlr-v119-koh20a} that predicts a label $y$ from an input $x$ through an intermediate ``concept'' prediction layer (Figure adapted from \citet{pmlr-v119-koh20a}). Panel (b) shows our proposal: after predicting concepts, the system interactively queries the human for true values $c_i$ for concepts chosen so as to maximize prediction accuracy and minimize acquisition cost.}
    \label{fig:overview}
\end{figure}

Deep learning-based AI systems have demonstrated significant capabilities across a range of applications. However, in many sensitive or safety-critical applications like healthcare or toxicity detection, AI-based predictive systems are seldom deployed in a standalone fashion; instead, they are used as one component in the overall decision-making workflow \citep{lee2021human}. 

\begin{figure*}[!ht]
    \centering
    \includegraphics[width=.9\linewidth]{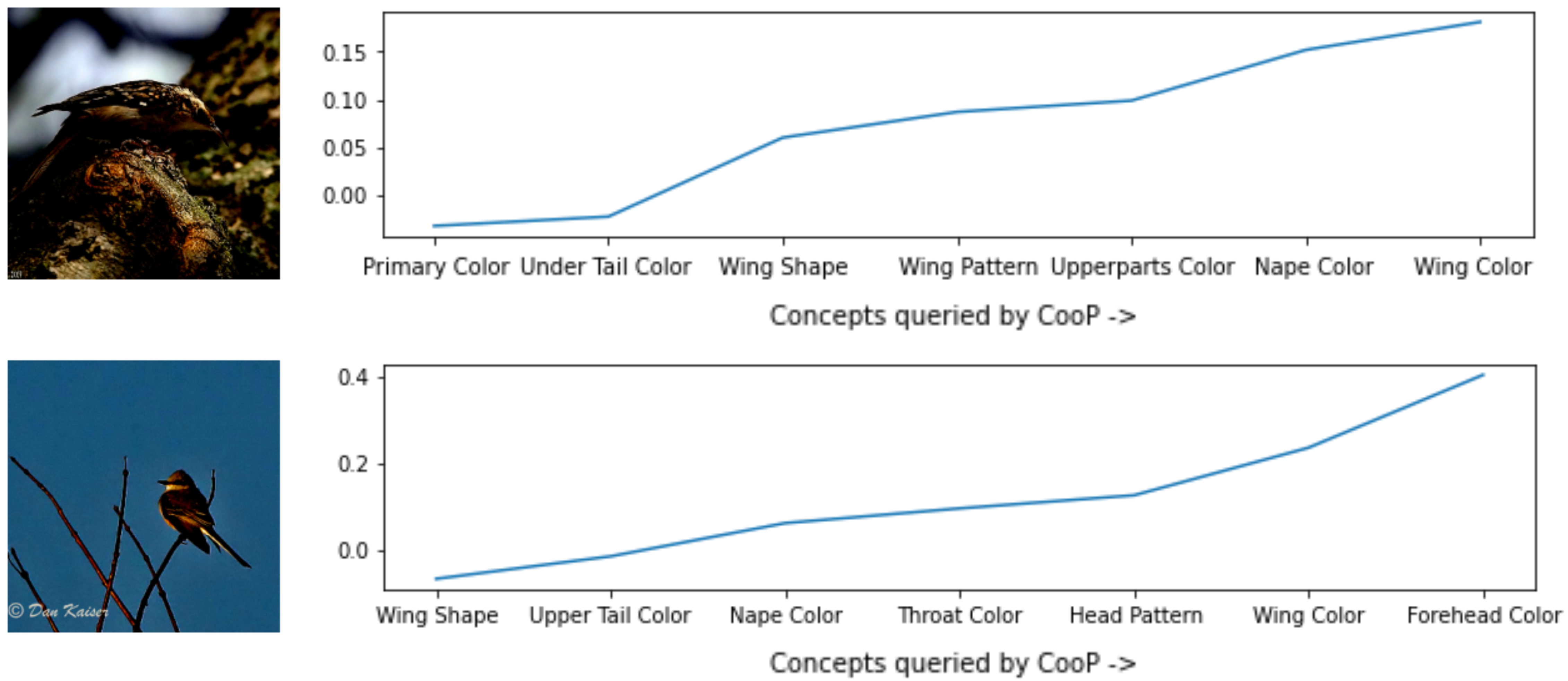}
    \caption{Influence of interventions using queries from the \ours\ policy on a \cub\ trained model on two example images. Y-axis denotes the difference between the probability assigned to the correct and top incorrect class.}
    \label{fig:coop_exemplars}
\end{figure*}

A concrete interactive setting that has received significant attention in literature is that of active feature acquisition (AFA) \citep{greiner2002learning, kanani2008prediction}. In this setting, a classifier can request additional features at prediction time. There is a cost associated with each feature acquisition, so the AFA algorithms have to reason about the value of the acquired feature relative to the cost. In this paper, we focus on a slightly different setting where the classifier always has access to a basic set of features (like pixels of an image). However, at prediction time, the classifier can request additional labels corresponding to human-interpretable concepts that are relevant to the prediction task. For example, when predicting bird species from a bird image, the classifier can request access to the wing shape. A key distinguishing feature of this framework from the general active feature acquisition setting is that the human-interpretable concepts are potentially inferrable from the initial features input to the model. In particular, we build on Concept bottleneck models \citep{pmlr-v119-koh20a} which explicitly predict concept labels from images and then predict the final label based on the concept label predictions (Figure~\ref{fig:overview}(a)). The authors argue that CBMs show better explainability, performance under distribution shift, and improvements in performance when concept labels are made available at prediction time. However, they only consider static policies for test time intervention that request concept labels in a predetermined order.

We present Cooperative Prediction (\ours), a dynamic interactive policy that can reason about the uncertainty associated with a given test instance, and request only those concepts that improve predictive power on that instance. Thus, for a given number of queried concepts (or, more generally, a budget for concept acquisition cost), our interactive models can achieve significantly higher levels of performance than static baselines.  Figure~\ref{fig:coop_exemplars} illustrates with a pair of example images how our approach selects different sequences of concepts to be queried from the user, based on the ambiguity inherent in the specific instances, in a bird species identification task~\cite{wah2011caltech}. The specific sequence of queries allows us to rapidly improve our confidence in the true label.

\noindent We make the following \textbf{contributions}\footnote{Code is available at \url{https://github.com/google-research/google_research/tree/master/interactive_cbms}}:
\begin{itemize}
    \item We develop a simple approach for training policies that act on top of Concept Bottleneck Models (CBMs) from \citet{pmlr-v119-koh20a}, with the objective of achieving a specified trade-off between interaction cost and predictive performance. Our approach only has a couple of tunable parameters and can be learned using a small validation set separate from the training set used for the CBM.
\item We compare our dynamic instance-based query policy against static policies that determine a fixed order of querying attributes for all examples \citep{pmlr-v119-koh20a}, as well as SOTA active feature acquisition strategies \citep{jointafa} that apply to more general settings of acquiring features beyond human-interpretable concepts.
\item Our model incorporates and optimizes for a cost model of feature acquisition; we show that our approach can adapt to settings with non-uniform costs of querying attributes, and demonstrate superior performance compared to the baselines.
\end{itemize}

% \subsection{Datasets}

% We evaluate our approach on several datasets consisting of a high dimensional input (images or text) paired with categorical attributes that can be queried upon request from the model. In particular, we use the Birds and OAI datasets from \citet{pmlr-v119-koh20a}

\section{Related Work}
The closest relevant work to our paper is that on Active Feature Acquisition and Concept-Aware Models. We review the literature in both and explain how our work is distinguished from this prior work:

\textbf{Active Feature Acquisition (AFA):}
The goal of active feature acquisition is to acquire a subset of features, in a cost sensitive manner, for each instance in order to maximize performance at test time. \citet{zubek2004pruning} proposes a AO* based learning algorithm to heuristically search for classification policy. \citet{ma2018eddi} and \citet{zannone2019odin} uses a partial variational autoencoder to predict the rest of features given the acquired ones to model the feature importance and uncertainity and combine it with acquisition policy to maximize information gain. \citet{jointafa} treats this as a joint learning problem trains both the classifier and RL agent together to learn when and which feature to acquire to increase classification accuracy while maintaining cost-efficiency. \citet{li2021active} reformulate Markov Decision Process (MDP) and learn a generative surrogate model to capture inter feature dependencies to aid RL agent with intermediate rewards and auxiliary information.

In this paper, we deal with a special case of the general AFA problem where the classifier always has access to an initial set of features (like the pixels in an image), but can request additional human-interpretable concept labels for each prediction. The goal of our work, just like in AFA, is to select which concepts to acquire labels for in a cost sensitive manner. However, we leverage the fact that we are not acquiring arbitrary features but human-intepretable concepts that have strong correlations to the input features and the final label, and exploit the fact that we can train a CBM to solve the prediction task. This makes our approach more perfomant than the general AFA approach. 

In \citet{branson2010visual}, the authors work in exactly the same setting as us and even obtain results on one of the datasets we use. In this work, the authors posit a generative model for the concept labels given the final label, and assume that the concepts are independent of the input features given the final label. They exploit this assumption to compute a tractable posterior distribution on the label given additional concept labels acquired. Our work does not make any such assumptions, instead relying on the CBM to learn the correlations between the input features and concepts, and the influence of the concept labels on the final label. We achieve stronger results empirically than the approach presented in \citet{branson2010visual}.

\textbf{Concept Aware Models:} Concept bottleneck models \citep{pmlr-v119-koh20a} were developed to show that building models that explicitly predict concept labels from images or other raw features helps with explainability, performance under distribution shift and improvements in performance when concept labels are made available at prediction time.  CBMs have been extended in various ways: \citet{Bahadori2021} performs causal reasoning to debias CBMs, \citet{antognini2021rationalization} develops textual rationalizations based on concept interventions.   However, of these works, only \citet{pmlr-v119-koh20a} explicitly considers test time intervention, and even here they only consider static policies for test time intervention that request concept labels in a fixed predetermined order (learned on a validation set). In this work, we allow for dynamic interactive policies that, for each prediction, reason about which concept labels are useful to improve the reliability of the prediction and request only those. Thus, with the same number of concepts allowed to be queried at prediction time, our interactive models can achieve significantly higher levels of performance than static baselines.\\

\section{Methods}

We outline the problem formulation as well as a simple approach to computing interactive policies on top of pretrained predictive models.

\subsection{Formalizing an Interactive Prediction System}

\emph{Input, Concept and Label Spaces:} We denote raw input features by $x$ (these can correspond, for example to images or text or features derived from these), the labels $y$ and the acquired concepts $c$. We assume that $c$ is a vector of $m$ categorical concepts and $y$ is a categorical scalar taking $K$ possible values. Individual concepts are denoted $c_i, i=1, \ldots, m$. The set of possible values $c$ can take is denoted $\mathcal{C}$ (with the set of possible values for $c_i$ being denoted $\mathcal{C}_i=\{1, \ldots, n_i\}$ so that $\mathcal{C}=\prod_i \mathcal{C}_i$) and the set of possible values $y$ can take is denoted $\mathcal{Y}=\{1, 2, \ldots, K\}$.  We denote the space of possible inputs by $\mathcal{X}$. We also note here that the term concept is loosely applied - in some contexts, it may refer simply to additional attributes (a user's age or gender, for example) or pieces of information that can be acquired at some cost at prediction time. Hence we use the term concepts or attributes interchangeably.

\emph{Concept Bottleneck Model:} The CBM is the composition of an input-to-concept (\ftoc) model $p_\theta(c|x)$ and a concept-to-label (\ctoy) model $p_\phi(y|c)$. We assume that both models make probabilistic predictions and that the input-to-concept model makes an independent probabilistic prediction for each concept $c_i$, denoted by $p_\theta(c_i|x)$ for each $i=1,\ldots, m$. 
We assume that these models have been trained and are available to us and do not make any assumptions about how the models were trained, or whether the predictions output represent calibrated probabilities.

\emph{Intervention:} In the absence of interactivity, the two stage model makes predictions by first invoking the \ftoc~ model and then passing the output to the \ctoy~ model to get the final prediction. However, in our setting, we allow for interventions on the concepts, i.e, replacing the predicted value (or distribution over values) of a concept with its ground truth value. We denote the prediction concept values as $\hat{c}=p_\theta\br{\cdot|x}$ and the intervened concept values (i.e. the ground truth) as  $c$. We further denote the prediction of a \ctoy model with partial intervention on a subset of concepts $S$ as $p_\phi\br{y|c_S, \hat{c}_{\Sc}}$. We also use the notation $p_\phi\br{c_S=v, \hat{c}_{\Sc}}$ to denote the predictions given the intervention where the concepts $c_S$ are set to a specific value $v$.

\emph{Interaction Cost Model:} We denote the cost of acquiring an attribute $c_i$ as $\cost_i > 0$. We assume all costs are positive and that the cost of acquiring a concept is the same independent of the previously acquired concepts or the value the concept takes. The total cost of acquiring a set of attributes $S$ is assumed to be $\sum_{i \in S} \cost_i$. Extensions that don't make these assumptions are possible but we leave this to future work. We assume that for each prediction made, there is a budget $B$ and that the interaction can only occur while the total cost of concepts acquired so far is below $B$.

\emph{Interactive Policies:} Given the two stage model, we define an interactive policy $\pol$ as follows: An interactive policy takes as input a set of revealed concepts $c_S$ where $S \subseteq \{1, \ldots, m\}$, the \ftoc~ and \ctoy~ models and interaction costs $q$ and outputs the new concept to acquire:
\[\pol\br{S, p_\theta, p_\phi, q, B} = i \in \Sc = \{1, \ldots, m\} \setminus S\]
We will usually drop the dependence on $p_\theta, p_\phi, q$ (as these are assumed to be always available) and simply write $\pol\br{S}$. A rollout of an interaction policy corresponds to the Algorithm \ref{alg:PolicyRollout}. The final prediction generated on an input $x$ is denoted $\rollout\br{\psi, x}$

\begin{algorithm}
\begin{algorithmic}
\State $S \gets \emptyset$ 
\While {$b \leq B$}
    \State $i \gets \pol\br{S}$ 
    \If {$b \leq B-q_i$}
        \State Acquire the label for concept $c_i$ 
        \State $S \gets S \cup \{i\}$
        \State $b \gets b + q_i$
    \Else
        \State $b \gets B + 1$
    \EndIf
\EndWhile
\State Output prediction $\argmax_{y \in \mathcal{Y}} p_\phi\br{c_S, p_\theta\br{c_{\Sc}|x}}$
\end{algorithmic}\caption{Policy Rollout}\label{alg:PolicyRollout}
\end{algorithm}

%Let us assume the concepts c are drawn from a set C, and that at any time we have observed some subset S of concepts. We want an iterative querying policy pi(x,c_S, \hat{c}_{\S}) = argmax_i crossent(p_\phi([c_{S union i};\hat{c}_{\S,i}. In other words, we start with the output of p_theta (i.e., estimates \hat{c}, then loop over the currently unseen concepts iteratively selecting the concept to next pick. etc.

% Cost version of teh same problem.

\emph{Separation of Policy and Model Learning:} In this paper, we restrict ourselves to learning an interactive policy as a post-hoc step on top of a learned model. We do not make assumptions about how the two-stage model (i.e., $p_\theta(\cdot), p_\phi(\cdot)$) is trained.

\emph{Dataset for Policy Learning:} We assume that we can acquire a dataset consisting of $(x, c, y)$ triplets sampled iid from an underlying unknown joint distribution $\Pdata$. Our goal is then to minimize
\begin{align}
    \ExP{\br{x, c, y} \sim \Pdata}{\ell\br{y, \rollout\br{\psi, x}}} \label{eq:InteractOpt}
\end{align}
where $\ell$ is a loss function that measures the discrepancy between the true label $y$ and the label output by rolling out the policy $\psi$. 
% The ideas developed in  \citet{pmlr-v119-koh20a} are one way to realize such models, but we do not make any assumptions about the training procedure for these models. Instead, we focus on the problem that given a validation set consisting of images, concept labels and the final prediction task label for each datapoint, how does one train an interactive policy that will act on top of pretrained models but enhance their peroformance by selective asking for values of certain concept labels for each test instance. 

%\blueps{

%We define the following optimization problem: (given concepts, costs, model predictions \& uncertainty, etc; learn a policy \pi that asks for concept labels in a particular order. 

%}

\subsection{Interactive Policy Learning with Cooperative Prediction (\ours)}
\label{sec:policies}
%\blueps{
%\todo{idk why the first letter 'C' in \cub~, \chexpert~ and \ours~ looks kinda weird}
We present Cooperative Prediction (\ours), a lightweight approach to learning interactive policies that attempt to optimize the objective in equation \eqref{eq:InteractOpt}. 

The key intuition behind \ours~ is that deciding which concept labels to acquire should be informed by three considerations: a) The uncertainty associated with the concept prediction - if we can already infer the concept label with high confidence based on the input features, there is not much value to acquiring it. b) The impact of the concept label on the final label prediction - If knowing the value of the concept does not change the predicted label confidence scores by much, it is not very valuable. c) Cost of acquiring the concept. 

\ours~ uses a very simple measure of each of the three components, and chooses concepts to acquire iteratively in a greedy fashion by developing a score function based on the three components and choosing the concept not yet acquired that scores the highest. In particular we use the following measures:

\begin{itemize}
    \item\emph{Concept prediction uncertainty (CPU):} We compute the entropy of the distribution $p_\theta\br{c_i|x}$ for each concept $c_i$ with $i \not \in S$
    \[\mathcal{H}\left[p_{\theta}(c_i|x)\right]\]
    \item\emph{Concept importance score:} We compute the expected change in the softmax score $p_\phi\br{y|c}$ associated with the final label prediction when the concept label for each concept $c_i$ is changed in the inputs to the \ctoy\ model, i.e.:
    \begin{align*}
        & \text{CIS}(c_i;S, k)  = \\ 
        & \left|\ExP{}{p_{\phi}\br{y=k|c_i=v, c_S, \hat{c}_{\Sc\setminus\{j \}}}}-p_{\phi}\br{y=k|c_S, \hat{c}_{\Sc}} \right|
    \end{align*}
    where $k$ was the label predicted in the previous round of the interaction and the expectation is taken over concept values $v \sim p_{\theta}(c_i|x)$.
    \item\emph{Acquisition cost:} This is simply the acquisition cost of each attribute $\cost_i$.
\end{itemize}
The final score is simply a linear combination of normalized versions of these scores (each score is normalized so the range of values it takes on the policy learning dataset lies between 0 and 1) and the overall attribute selection algorithm is outlined in algorithm \ref{alg:Selection}. Each linear combination of scores leads to a different interactive policy, and those weights are tuned to optimize performance on a holdout validation set.

\begin{algorithm}
\begin{algorithmic}
\State Given $p_\theta, p_\phi, q$, set of concepts acquired so far $S$ and highest scoring predicted label based on acquired concepts $k \in \mathcal{Y}$, and score importance weights $\alpha, \beta, \gamma \in \R_+$
\ForAll {$i \in \Sc$}
    \State Compute $\text{score}_i=\alpha \text{CPU}(c_i;S)  + \beta \text{CIS}(c_i;S, k) - \gamma \cost_i$
\EndFor
\State Output $\argmax_i \text{score}_i$
\end{algorithmic} \caption{\ours~ policy}\label{alg:Selection}
\end{algorithm}

%To design an interactive policy, we look at two things - concept prediction uncertainty and perceived concept importance for the final label prediction. We use the entropy of the individual concept predictions to quantify the uncertainty in the concept prediction.

%In the case of the CUB dataset, where we intervene on \textit{concept groups}, we use the sum of the entropies of the individual concept predictions that belong to a particular group. To measure the perceived importance of a given concept, we use the expected change in the probability of the predicted class upon observation of the concept, with the expectation computed under the concept model's predicted probabilities. For a given image $x_i$, and a concept (or concept group, with a slight abuse of notation) $c_j$ we define a score:
%$$
%\text{Score}_{x_i}^k = \mathcal{H}_{p_{\theta}}(c|x_i) + \alpha  \mathbb{E}_{c \sim p_{\theta}(c_j|x_i)} p_{\phi}(y=k|c_j=c)
%$$
%Where $k = \operatorname{argmax}_k p_{\theta, \phi}(y=k|x_i)$. We tune the mixing parameter $\alpha$ on the validation set.

%\todo{heuristic stopping criterion}
A primary advantage of the proposal is its ease of learning, especially in sparse-data scenarios, as it only requires that we estimate two mixing parameters. As the first paper proposing this novel problem setting (to our knowledge), our primary goal here is to demonstrate that careful policy selection can indeed provide value. We leave further improvements and theoretically principled approaches to this problem for future work, here instead focusing on demonstrating that a simple approach works well for this novel problem setting and formulation.

%We note that there are several formal treatments of the policy learning problem that are applicable here; for instance (XX,YY,ZZ with citations). In particular, an approach that has promise in the settings of sufficient data is to learn a nueral policy that is additionally condition on a \textit{feature representation of the input}, allowing the policy to leverage latent information not completely covered by the concept bottleneck alone.
%}

\section{Experiments}
\subsection{Datasets}

We experimented with the following datasets, with characteristics summarized in Table~\ref{tab:datasets}: \\
  \textbf{\cub\ (Caltech-UCSD Birds): } This dataset contains pictures of birds coupled with human-labeled concept attributes identifying prominent characteristics (wing color, beak length, undertail color, etc.) \citep{wah2011caltech}. The there are 28 such categorical concepts, resulting in a total of 112 binary labels. In an interactive setting, attributes are revealed at prediction time only when the policy queries an attribute. In practice, this could be seen as asking human labelers in an interactive setting to provide specific hints on concepts they can easily identify (like beak length or wing color) even if they are unable to make the final prediction on what species of bird it is, as most labelers will be unable to do this unless they are specifically trained on this task. \\
   \textbf{\chexpert: } This dataset contains chest X-rays accompanied by binary concept labels extracted from a report generated by a radiologist, with the goal of predicting whether the X-ray was normal or abnormal \citep{irvin2019chexpert}. Each chest X-ray is also accompanied by 13 binary attributes that include concepts easily recognized by a non-expert (presence of a fracture or any support devices on the patient), harder-to-label attributes that need a nurse or physician (e.g., Cardiomegaly), and, finally, attributes that require a radiologist to label. \\
   \textbf{\oai\ (Osteoarthritis Initiative):} This dataset contains knee X-rays, annotated with the Kellgren-Lawrence grade (KLG), a 4-level ordinal variable (assumed to be categorical for training) that measures the severity of knee osteoarthritis. Each knee X-ray is also annotated with 10 ordinal attributes describing joint space narrowing, bone spurs, calcification, etc., resulting in a total of 40 binary concepts.

% \todo{include table with dataset details: following columns -- name, #input dims, #concepts, #train, #val,#test}

\subsection{Base Models}

Following the proposals made by~\citet{pmlr-v119-koh20a}, we train the following 2 kinds of concept-bottleneck models (CBMs):\footnote{ Although  \ours\ is agnostic to the specific way the base CBM has been trained, each base CBM may have idiosyncrasies in its predictive power that interact with \ours; therefore, specific base CBMs combined with \ours\ may perform overall better on any given dataset. }

\noindent \textbf{Independent model:} The \ftoc\ and \ctoy\ models are trained separately, respectively mapping the inputs $x$ to the true concepts $c$, and the true concepts $c$ to the labels $y$ respectively. The \ctoy\ model sees true concept values as input at train-time, and estimated values (probabilities) at test time. \\
\noindent \textbf{Joint model:} Both \ftoc\ and \ctoy\ models are learned using a joint optimization criterion which combines the concept prediction loss (cross-entropy) and label prediction loss (cross-entropy). The probabilities output by the \ftoc\ model are passed on to the \ctoy\ model during training. \\
% \noindent \textbf{Joint sigmoid model:} Similar to the Joint model above, except that the concept prediction outputs (logits) from the \ftoc\ models are converted to probabilities using a sigmoid function.

 We build our interactive intervention models on top of these CBMs, and propose and evaluate various methods of intervention using each of these as the base CBM. Although we performed extensive experimentation with both the Independent and Joint models, due to space considerations we present only results from the Independent CBM here. Findings on the Joint model are qualitatively similar; please refer Appendix \ref{sec:ablation} for more details.

\subsection{Training and Evaluation}

For each experiment, we split the data into 3 sets: train, validation, and test -- the details are available in Table~\ref{tab:datasets}. We used the training data to train the base CBMs, and validation data to select parameter settings, if any, for intervention policies. We then retrained the base model on pooled train + validation datasets\footnote{We did this due to data paucity in the datasets we studied.  This is common practice for the datasets, and does not compromise the findings since all reported results are on an unseen test set.}, and finally reported performance of the policies on the unseen test set. In case an intervention policy did not require parameter selection, we directly trained the base model on the pooled train+val data. Finally, for \ours, we require accurate measures of uncertainty in order to make effective decisions; we, therefore, calibrate the pooled concept probabilities across the training data for a given base model using isotonic regression.

Our primary metric is accuracy for \cub\ and \oai, and AUC for \chexpert\ at the classification tasks. For each intervention policy, we start with performance using only the input $x$. We then iteratively obtain true labels for intermediate concepts $c$ as specified by the policy, set the concept value to the observed true value, and report the performance of the updated prediction after adding the observed concept. In this manner, we obtain a curve measuring the performance metric as a function of the number of observed concepts.

% \begin{table}[!h]
% \fontsize{9pt}{10pt} \selectfont
% \centering
% \begin{tabular}{l|c|c|c|c|c}

% Dataset Name       & Input Dims  & Concepts & Train & Val & Test                \\ \hline
% \cub    & $299 \times 299 \times 3$ & 112 & 4796  & 1198  & 5794\\
% \chexpert  &  $320 \times 320 \times 3$  & 13 & 178731 & 22341 & 22342\\
% \midrule \midrule
% \end{tabular}
% \caption{Datasets used in our experiments, and associated statistics. }\label{tab:datasets}
% \end{table}
%\todo{complete the data table}

\begin{table}[!ht]
\fontsize{9pt}{10pt} \selectfont
\centering
\begin{tabular}{l|c|c|c}

      & \cub  & \chexpert & \oai \\ 
\midrule
Input Dims    & (299, 299, 3) & (320, 320, 3) & (512, 512, 3)\\
Concepts  & 112  & 13 & 10 \\

Data splits & & & \\ 
~~train  & 4,796  & 178,731 & 31,370 \\
~~val  & 1,198 & 22,341 & 4,426 \\
~~test  & 5,794  & 22,342 & 4,522\\
% \midrule
\end{tabular}
\caption{Details of the datasets used in our experiments.}\label{tab:datasets}
\end{table}

%\redps{We present results for the joint CBM model (qualitatively similar) in the supplementary materials.}
%Also, comparing the three panels, the same relative ordering is confirmed across different \cbm\ architectures, although the chosen information mixing parameter varies.

% \todo{any comparative differences e.g., on overall accuracy and/or interaction with various policies}
%\todo{complete the figure captions}
% \begin{figure}[!th]
%     \centering
%     \includegraphics[width=\linewidth]{Figures/InterventionPolicy_gains.png}
%     \caption{Accuracy gains vs interaction cost (unit cost model) on \cub. See text for details. }
%     \label{fig:intervention_cub}
% \end{figure}

% \begin{figure}[!th]
%     \centering
%     \includegraphics[width=\linewidth]{Figures/InterventionPolicy_Chexpert.png}
%     \caption{Accuracy gains vs interaction cost (unit cost model) on \chexpert.  See text for details. }
%     \label{fig:intervention_chexpert}
% \end{figure}

\begin{figure*}[!th]
    \centering
    \includegraphics[width=0.32\linewidth]{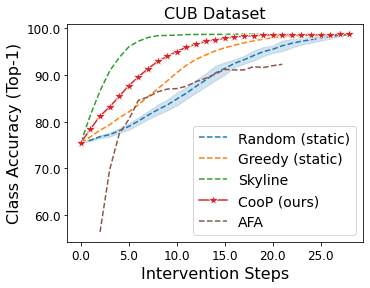}
    \includegraphics[width=0.32\linewidth]{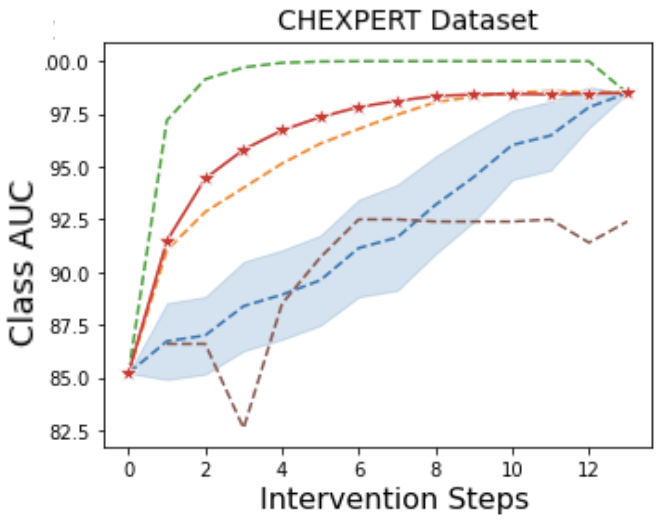}
    \includegraphics[width=0.32\linewidth]{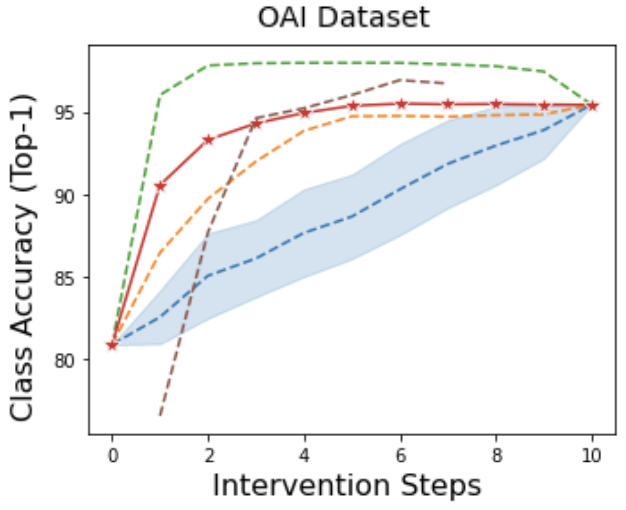}
    \caption{Accuracy gains vs interaction cost (unit cost model) on different datasets.  See text for details. }
    \label{fig:interventions}
\end{figure*}

\subsection{Baselines for Comparison}
\textbf{\greedy:} Select an ordering of attributes using a greedy ranking scheme over the validation dataset; i.e., the first attribute in the list is the one that improves the performance measure the most on average over the validation set. Subsequent attribute orders are determined in a similar greedy fashion after conditioning on all previous attributes as being available.  \\
\textbf{\random:} For each instance in the training set, choose the next attribute to query at random. \\
\textbf{\skyline:} Evaluate an oracle greedy approach which checks, for each instance in the \textit{test set}, the specific greedy order of querying attributes that provides maximum incremental gains on each step for that test instance. This is an \textit{oracular} skyline since it uses the test label for optimization, and is an approximate \textit{ceiling} on the performance achievable under any interactive policy.\footnote{The reason this skyline is only approximate-oracular is that searching all possible orderings of attributes for querying is combinatorially infeasible; as seen in our results, the proposed greedy approximation quickly approaches 100\% performance measure suggesting it is a good approximation.}\\
\textbf{Active Feature Acquisition (\AFA):} We also compare against the Active Feature Acquisition policy based on work by \citet{jointafa}. We use image embeddings from ResNet18 pre-trained on ImageNet as auxiliary information, and ground truth concepts as features that can be acquired; we train an RL policy to actively acquire features/concepts as described in \citet{jointafa}. This algorithm is relatively sensitive to the value of a hyperparameter $r\_cost$ that determines the tradeoff between model performance and acquisition cost. For evaluating \AFA, we performed a hyperparameter search for $r\_cost$ in the range of [-1e-5, -0.06] for each number of intervention steps using the accuracy/AUC on the validation set. We also tried using fine-tuned ResNet features instead of pre-trained features as auxiliary information, but it resulted in severe overfitting on training data. For a fair comparison, we use the same data splits as other baselines and the policy is trained until convergence.

% \begin{figure*}[!ht]
%     \centering
%     % \includegraphics[width=.75\linewidth]{LaTeX/Figures/cub_cost_intpolicy_both.png}
%     % \includegraphics[width=.75\linewidth]{LaTeX/Figures/chexpert_cost_intpolicyv2_both.png}
%     \includegraphics[width=0.3\linewidth]{LaTeX/Figures/cub_intpolicy_wcost.png}
%     \includegraphics[width=0.3\linewidth]{LaTeX/Figures/chexpert_intpolicy_wcost.png}
%     \caption{Cost-efficient interventions on different datasets. For \cub, we report mean results for 10 random cost assignments.}
%     \label{fig:costmodel}
% \end{figure*}

\begin{figure*}[!ht]
    \centering
    \includegraphics[width=0.32\linewidth]{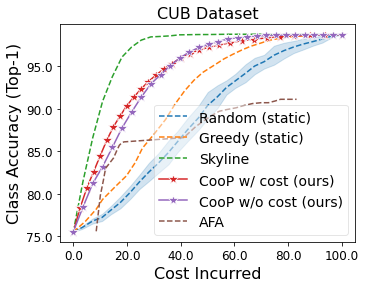}
    \includegraphics[width=0.32\linewidth]{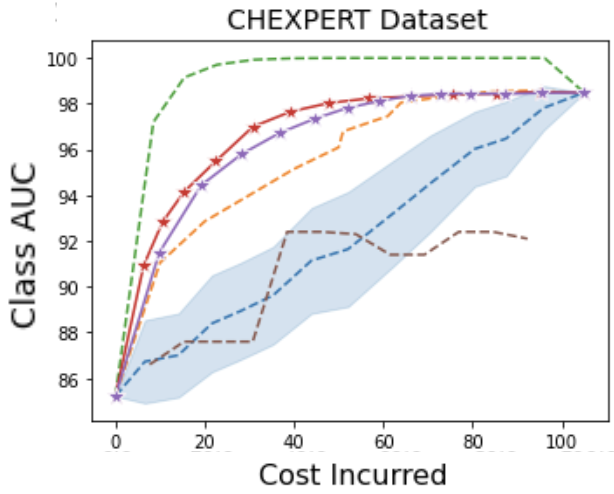}
    \includegraphics[width=0.32\linewidth]{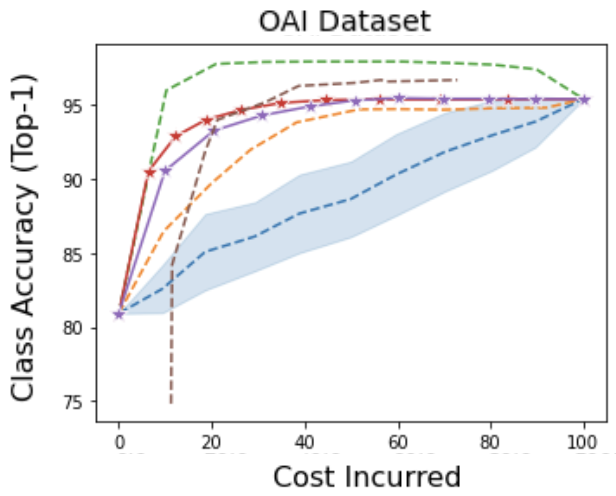}
    \caption{Cost-efficient interventions on different datasets. For \cub\ and \oai, we report mean results for 10 random cost assignments. For \chexpert, we use the domain-informed cost model.}
    \label{fig:costmodel}
\end{figure*}

\subsection{Intervention Costs}
\label{sec:costing}
In practice, it is likely that obtaining certain concept labels has a higher cost, for example, due to the difficulty of the annotation task or the data being purchased at a cost from a data provider, or the privacy costs of asking for sensitive user information.  We studied the following cost models: \\
\emph{Unit cost}: Each query made by the interactive policy incurs a unit cost. This is largely appropriate for the \cub~dataset since the concepts are largely identifiable easily by a non-expert. \\
\emph{Random cost}: To stress test our approach, we also experiment with random cost models where a randomly chosen cost from the range $[1, 7]$ is assigned to each concept, which is then normalized such that the total cost of all concepts is 100.\\ % \todo{I've changed the cost here. I'm using integer costs sampled uniformly from the range [1,6], both inclusive}  \\
\emph{Systematic cost}: In datasets like \chexpert~, it is clear that some attributes can be easily labeled by non-experts while others require a specialized radiologist. Thus we assign systematic costs, that have a strong justification depending on the difficulty of acquisition of a concept label. Based on consultations with domain experts, we use concept acquisition costs of 1, 3, and 10 for concepts that are very easy, moderately difficult, and very difficult to annotate respectively. Similar to random costs, we also normalize \chexpert's systematic costs such that the total cost of all concepts is 100.

The costs are factored into the learning of policies for \ours\ as described in Section~\ref{sec:policies}.  For the \chexpert\ dataset where a systematic cost was available, we used it to evaluate our cost optimization procedure; for \cub\ and \oai, we used random costs.

\section{Results}

We perform experiments that demonstrate the following valuable contributions from \ours~:\\
\textbf{Performance gains from adaptive interactive policies:} Section~\ref{sec:results_perf} shows that by querying just 5 additional concept labels at prediction time, our interactive policies can improve relevant performance metric (AUC or accuracy) by 20-25\% over any static baseline that determines in advance an order in which attributes are queried. In some cases, \ours~ achieves a sizable fraction of the possible gain determined by the skyline. Improving \ours~ further to fully close the gap to the skyline is a compelling direction for future work.\\
%\textbf{Data minimization achieved by adaptive interactive policies}: Since it is clear that having accurate concept labels can help predictive performance, a model that always has access to the concept labels should perform better than any interactive policy. However, we show that interactive adaptive policies can achieve data minimization, i.e, use less data than these models by only querying attributes that are necessary for a prediction task (Section~\ref{sec:results_minim}). \ours~ can reduce test-time data usage by 50\% while incurring only 5\% performance drop relative to a model that uses all concept labels.\\
\textbf{Cost-aware acquisition}: Relative to policies that only use uncertainty to drive interactivity,  \ours~ is cost-aware and acquires concept labels only when they are valuable, achieving a better tradeoff between cost and performance than simpler policies (Section~\ref{sec:results_cost}). \\
We conducted additional analyses, including sample efficiency for \ours, and ablation studies probing the contributions of different factors in our scoring function. For details refer to Appendix \ref{sec:ablation}.

\begin{figure*}[!ht]
    \centering
    \includegraphics[width=.9\linewidth]{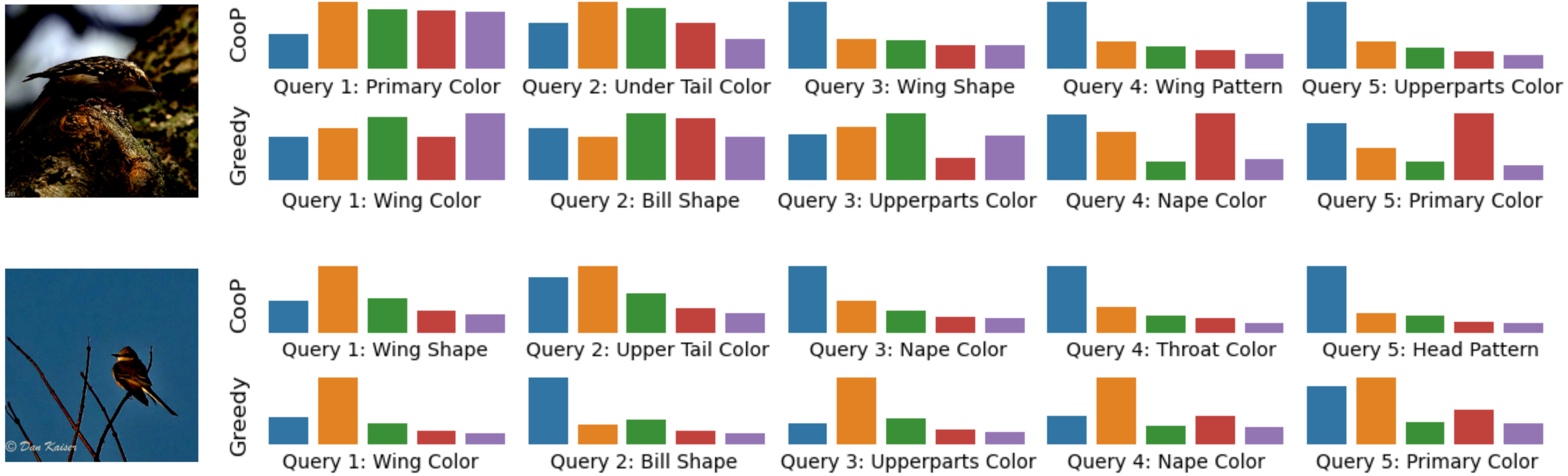}
    \caption{Comparison of the behavior of \ours\ and \greedy\ policies for the first five steps of intervention using two example images. The bar plots show the probabilities assigned to top 5 ranking classes according to the model, with the first blue bar representing the correct class. The bar plot titles represent the concept revealed by the respective policy.}
    \label{fig:coop_greedy_exemplars}
\end{figure*} 

\subsection{Predictive Performance Improvement}
\label{sec:results_perf}

Figure~\ref{fig:interventions} shows the results of our experiments on three different datasets: \cub, \chexpert, and \oai. 

Across all 3 datasets, we see that the simple \greedy\ policy already outperforms \random\ consistently across the range of intervention steps. This serves as a strong existence proof of nontrivial intervention policies. Note that the greedy policy requests concept labels in the same sequence regardless of the test data instance. In contrast, \ours\ uses instance-specific uncertainties both in concept labels and final predictions, and as a result significantly outperforms \greedy, again across datasets. In particular, \ours\ is able to substantially improve the performance metric with as few as 5 queries.

We also note that the \AFA\ baseline has an uneven performance across datasets and the range of intervention steps. For instance, the initial performance of AFA is quite poor, followed by a rapid rise in the \cub\ and \oai\ datasets. In \chexpert, \AFA\ fails to improve upon \random, a weak baseline. Finally, on the \oai\ dataset, \AFA\ does outperform \ours, although only by small margins and after the key first few queries. This variable performance is driven by two factors: the significant data need for the \AFA\ algorithm, and the need for selecting a particular tradeoff cost at training time, rather than being able to smoothly adjust the cost-benefit tradeoffs at test time on a per-instance basis. 

% Finally, we note that the choice of CBM (Independent vs Joint) may also influence the data efficiency of the test-time intervention: \ours\, as well as baselines, appear to use interventions more efficiently in the Independent model.

%\begin{figure}[!th]
%    \centering
%    \includegraphics[width=0.75\linewidth]{LaTeX/Figures/DataMinimizationInd.png}
%    \caption{Data minimization capabilities of \ours, demonstrated via drop in test accuracy on \cub\ relative to a \ctoy model that has access to ground truth concept labels on every example. The x-axis shows the number of interaction steps. Results demonstrate that using just 10-15 interactions, \ours~ can get within 5\% of the accuracy of a highly performant \ctoy~ model that achieves near-perfect accuracy on the \cub\ dataset.}
%    \label{fig:reldrop_cub}
%\end{figure}

% Figure~\ref{fig:interventions} (bottom row) replicates these results on the \chexpert\ dataset, with very similar findings for the different algorithms, and also across base CBM models. We note that the \AFA\ baseline performs significantly worse, and often below random-ordered intervention. This suggests that the \AFA\ training procedure is not able to align the learned rollout policy and the classifier, possibly due to the sparse \& delayed rewards in the setting used for training the policy.

An interesting finding is that for {all datasets}, \skyline\footnote{Although additional information should never worsen performance, we do see that adding certain concepts decreases accuracy, particularly in the Skyline. This is due to the heuristic nature of the Skyline, and the sparse-data settings that limit the base CBM's exposure to certain rare concepts in the training data.} performs noticeably better than both the other baselines and \ours.  Even though \skyline\ is an oracle, this finding suggests the possibility of additional headroom available for other sophisticated, potentially data-hungry policies.

Figure \ref{fig:coop_greedy_exemplars} illustrates \ours’s ability to customize intervention queries to each instance, which is its key differentiating feature as compared to \greedy. In the first image, \greedy\ chooses to reveal concepts like bill shape and wing color, which can be reasonably inferred from the image. \ours\ instead chooses to reveal under tail color which is difficult to see in the shadow, and the wing shape which can't be inferred since the bird's wings are closed. Similarly, in the second image, most concepts that \greedy\ chooses to reveal are visible to some extent. \ours\ instead queries concepts like upper tail color and head pattern which are hidden in the shadows, and wing shape, which also can't be inferred since the wings are closed.

%\subsection{Data minimization at test-time}
%\label{sec:results_minim}
%Figure~\ref{fig:costmodel} shows a different view of the data in the initial experiment (Figure~\ref{fig:intervention_cub})--here, we show the relative accuracy drop of the intervention model compared to a model that has all concept labels available at prediction time. We see that  \ours\ reduces the gap to the full-data model faster than \greedy\ for all three base CBM settings. These findings are important for settings with privacy considerations or sensitive data such as medical history, as \ours\ minimizes the need to access these concepts or attributes. An interesting finding is that the relative gap is reduced significantly faster in the Independent Model setting, particularly for \ours, when compared against the Joint and Joint Sigmoid settings. \ours\ also appears to converge to a smaller relative gap in the Independent model compared to the other settings; these findings suggest that the combination of the Independent CBM and \ours\ may provide the best gains. %\todo{Remark on chexpert as well, referencing the appendix}

\subsection{Cost-Efficient Interventions}
\label{sec:results_cost}

In the previous experiments, we assumed that all interventions (concept labels) have unit cost; we now explore the scenario where different concepts have different costs (see Section~\ref{sec:costing} for details on the cost model). Figure~\ref{fig:costmodel} shows the result of \ours\ when optimizing for intervention costs. The presented data is similar to Figure~\ref{fig:interventions}, except that the tradeoff is now accuracy versus total cost, as opposed to the total number of steps previously.   We see that \ours\ outperforms the baselines, both without and with cost-sensitive selection, and \ours~ with the cost is better.  This demonstrates \ours's ability to incorporate cost structure into the optimization of the interactive policy. 

% The results are similar on both datasets shown in the figure--note that for \cub\ we show an average over 10 random cost-assignment experiments due to the absence of a more naturalistic or domain-relevant relative cost for concepts. 

% \blueps{data cost of learning \ours\ policy: examining some summary metric of policy performance as a function of validation data size used in estimating parameters. Finding: we can learn \ours\ policies in a data-efficient manner.}
% \blueps{Scatterplot concept entropy vs average rank: do "skewed" / rare concepts get selected earlier, or later?}
% \blueps{Synthetic data experiments: examining how ease of concept prediction from input/ease of label prediction from concept each modulate the overall success of \ours-like policies.}

\section{Discussion}

We proposed a novel problem setting, that of iterative/interactive refinement of model predictions using human inputs, and the cost-efficient optimization of this interactive loop. We demonstrated a principled first-cut approach at learning such optimization policies in the context of two-stage, or \textit{concept-bottleneck} models where interactions are simplified to querying concept or attribute labels. We do not provide a wide or exhaustive discussion of the advanced algorithmic possibilities; indeed, we anticipate that future work will explore a number of alternate formulations both of learning the base models, and of optimizing interactive policies on top of those base models. In particular, one could hypothesize architectures other than the two-stage CBMs explored here. Further, human inputs could include information other than the bottleneck concepts -- for instance, side information that cannot be inferred from the input data, region-of-interest annotations, etc. A key related challenge for our work (and concept bottleneck models in general) is the need for intermediate supervision in the form of concept labels. Future work could explore weak or distant supervision for obtaining these concept labels.  Finally, there would be a need to evaluate our approach in realistic human-AI collaboration setups, where UX or other psychological factors may impact the interactive performance of the human-AI team \citep{bansal2021does}.

\section*{Ethics Statement}
Our goal is to increase the interpretability and robustness of predictive models, a significant net positive especially for applications such as medical diagnosis. As such, we do not expect any adverse outcomes of our work or follow-on research. For our experiments, we have used open-sourced datasets collected with appropriate review processes; we have not conducted human-in-the-loop experiments as it was out of scope for this paper. For an eventual system that includes humans in the predictive workflow, our proposal  only leverages additional instance specific data at test-time, as supplied by the expert responsible for prediction; this minimizes the potential for data misuse by our model.  

% \paragraph{Limitations \& Social Impact}
% Limitations were discussed in the previous section: proposing one approach among many,  dependence on intermediate supervision, and the need for online experiments to confirm the value. The primary application area of our work is in enabling interpretable, collaborative, and data-efficient predictions; these goals are primarily social goods, and may significantly mitigate the current risk of black-box ML models going awry, especially in mission-critical applications such as healthcare. %As such, we see this line of work as having the potential for positive social impact.

% Use \bibliography{yourbibfile} instead or the References section will not appear in your paper
\bibliography{aaai23}

\begin{thebibliography}{15}
\providecommand{\natexlab}[1]{#1}

\bibitem[{Antognini and Faltings(2021)}]{antognini2021rationalization}
Antognini, D.; and Faltings, B. 2021.
\newblock Rationalization through Concepts.
\newblock In \emph{Findings of the Association for Computational Linguistics:
  ACL-IJCNLP 2021}, 761--775.

\bibitem[{Bahadori and Heckerman(2021)}]{Bahadori2021}
Bahadori, M.~T.; and Heckerman, D.~E. 2021.
\newblock Debiasing concept-based explanations with causal analysis.
\newblock In \emph{ICLR 2021}.

\bibitem[{Bansal et~al.(2021)Bansal, Wu, Zhou, Fok, Nushi, Kamar, Ribeiro, and
  Weld}]{bansal2021does}
Bansal, G.; Wu, T.; Zhou, J.; Fok, R.; Nushi, B.; Kamar, E.; Ribeiro, M.~T.;
  and Weld, D. 2021.
\newblock Does the whole exceed its parts? the effect of ai explanations on
  complementary team performance.
\newblock In \emph{Proceedings of the 2021 CHI Conference on Human Factors in
  Computing Systems}, 1--16.

\bibitem[{Branson et~al.(2010)Branson, Wah, Schroff, Babenko, Welinder, Perona,
  and Belongie}]{branson2010visual}
Branson, S.; Wah, C.; Schroff, F.; Babenko, B.; Welinder, P.; Perona, P.; and
  Belongie, S. 2010.
\newblock Visual recognition with humans in the loop.
\newblock In \emph{European Conference on Computer Vision}, 438--451. Springer.

\bibitem[{Greiner, Grove, and Roth(2002)}]{greiner2002learning}
Greiner, R.; Grove, A.~J.; and Roth, D. 2002.
\newblock Learning cost-sensitive active classifiers.
\newblock \emph{Artificial Intelligence}, 139(2): 137--174.

\bibitem[{Irvin et~al.(2019)Irvin, Rajpurkar, Ko, Yu, Ciurea-Ilcus, Chute,
  Marklund, Haghgoo, Ball, Shpanskaya et~al.}]{irvin2019chexpert}
Irvin, J.; Rajpurkar, P.; Ko, M.; Yu, Y.; Ciurea-Ilcus, S.; Chute, C.;
  Marklund, H.; Haghgoo, B.; Ball, R.; Shpanskaya, K.; et~al. 2019.
\newblock Chexpert: A large chest radiograph dataset with uncertainty labels
  and expert comparison.
\newblock In \emph{Proceedings of the AAAI conference on artificial
  intelligence}, volume~33, 590--597.

\bibitem[{Kanani and Melville(2008)}]{kanani2008prediction}
Kanani, P.; and Melville, P. 2008.
\newblock Prediction-time active feature-value acquisition for cost-effective
  customer targeting.
\newblock \emph{Advances in neural information processing systems (NIPS)}.

\bibitem[{Koh et~al.(2020)Koh, Nguyen, Tang, Mussmann, Pierson, Kim, and
  Liang}]{pmlr-v119-koh20a}
Koh, P.~W.; Nguyen, T.; Tang, Y.~S.; Mussmann, S.; Pierson, E.; Kim, B.; and
  Liang, P. 2020.
\newblock Concept Bottleneck Models.
\newblock In III, H.~D.; and Singh, A., eds., \emph{Proceedings of the 37th
  International Conference on Machine Learning}, volume 119 of
  \emph{Proceedings of Machine Learning Research}, 5338--5348. PMLR.

\bibitem[{Lee et~al.(2021)Lee, Siewiorek, Smailagic, Bernardino, and
  Berm{\'u}dez~i Badia}]{lee2021human}
Lee, M.~H.; Siewiorek, D.~P.; Smailagic, A.; Bernardino, A.; and Berm{\'u}dez~i
  Badia, S. 2021.
\newblock A Human-AI Collaborative Approach for Clinical Decision Making on
  Rehabilitation Assessment.
\newblock In \emph{Proceedings of the 2021 CHI Conference on Human Factors in
  Computing Systems}, 1--14.

\bibitem[{Li and Oliva(2021)}]{li2021active}
Li, Y.; and Oliva, J. 2021.
\newblock Active feature acquisition with generative surrogate models.
\newblock In \emph{International Conference on Machine Learning}, 6450--6459.
  PMLR.

\bibitem[{Ma et~al.(2018)Ma, Tschiatschek, Palla, Hern{\'a}ndez-Lobato,
  Nowozin, and Zhang}]{ma2018eddi}
Ma, C.; Tschiatschek, S.; Palla, K.; Hern{\'a}ndez-Lobato, J.~M.; Nowozin, S.;
  and Zhang, C. 2018.
\newblock Eddi: Efficient dynamic discovery of high-value information with
  partial vae.
\newblock \emph{arXiv preprint arXiv:1809.11142}.

\bibitem[{Shim, Hwang, and Yang(2018)}]{jointafa}
Shim, H.; Hwang, S.~J.; and Yang, E. 2018.
\newblock Joint Active Feature Acquisition and Classification with
  Variable-Size Set Encoding.
\newblock In \emph{NeurIPS}, 1375--1385.

\bibitem[{Wah et~al.(2011)Wah, Branson, Welinder, Perona, and
  Belongie}]{wah2011caltech}
Wah, C.; Branson, S.; Welinder, P.; Perona, P.; and Belongie, S. 2011.
\newblock The caltech-ucsd birds-200-2011 dataset.
\newblock \emph{Computation \& Neural Systems Technical Report}.

\bibitem[{Zannone et~al.(2019)Zannone, Hern{\'a}ndez-Lobato, Zhang, and
  Palla}]{zannone2019odin}
Zannone, S.; Hern{\'a}ndez-Lobato, J.~M.; Zhang, C.; and Palla, K. 2019.
\newblock Odin: Optimal discovery of high-value information using model-based
  deep reinforcement learning.
\newblock In \emph{ICML Real-world Sequential Decision Making Workshop}.

\bibitem[{Zubek and Dietterich(2002)}]{zubek2004pruning}
Zubek, V.~B.; and Dietterich, T.~G. 2002.
\newblock Pruning Improves Heuristic Search for Cost-Sensitive Learning.
\newblock In \emph{Proceedings of the Nineteenth International Conference on
  Machine Learning}, ICML '02, 19–26. San Francisco, CA, USA: Morgan Kaufmann
  Publishers Inc.
\newblock ISBN 1558608737.

\end{thebibliography}
% \nobibliography{aaai23}

% \section{Acknowledgments}
% AAAI is especially grateful to Peter Patel Schneider for his work in implementing the original aaai.sty file, liberally using the ideas of other style hackers, including Barbara Beeton. We also acknowledge with thanks the work of George Ferguson for his guide to using the style and BibTeX files --- which has been incorporated into this document --- and Hans Guesgen, who provided several timely modifications, as well as the many others who have, from time to time, sent in suggestions on improvements to the AAAI style. We are especially grateful to Francisco Cruz, Marc Pujol-Gonzalez, and Mico Loretan for the improvements to the Bib\TeX{} and \LaTeX{} files made in 2020.

% The preparation of the \LaTeX{} and Bib\TeX{} files that implement these instructions was supported by Schlumberger Palo Alto Research, AT\&T Bell Laboratories, Morgan Kaufmann Publishers, The Live Oak Press, LLC, and AAAI Press. Bibliography style changes were added by Sunil Issar. \verb+\+pubnote was added by J. Scott Penberthy. George Ferguson added support for printing the AAAI copyright slug. Additional changes to aaai23.sty and aaai23.bst have been made by Francisco Cruz, Marc Pujol-Gonzalez, and Mico Loretan.

% \bigskip
% \noindent Thank you for reading these instructions carefully. We look forward to receiving your electronic files!

\appendix

\section{Dataset details}
\cub: \cub~ is publicly available and licensed under the CC-BY 4.0 license, permitting commercial and non-commercial work with attribution. We use the denoised version for the Caltech UCSD Birds dataset as used by~\citep{pmlr-v119-koh20a}. Due to noisy concept annotations, the concept labels for each image are aggregated into class-level concepts by replacing the instance-level concept labels with the majority vote for the class: e.g., if more than 50\% of crows have black wings in the data, then all crows are set to have black wings. Furthermore, the concepts that are too sparse (i.e. concepts that are activated only in less than 10 classes) are filtered out leaving us with a reduced set of 112 concepts.

\chexpert: We divide the standard training set of \chexpert~ \citep{irvin2019chexpert} into 80-10-10 train-val-test splits. Each chest radiograph in the dataset is labeled for the presence of 14 observations as positive, negative, or uncertain. For all 14 observations, we map the uncertain and negative labels to 0 and positive labels to 1, hence simplifying the task to binary classification. We use "No Finding" as the final label and use the remaining as concepts in the bottleneck. \chexpert~ is \href{https://stanfordmlgroup.github.io/competitions/chexpert/}{publicly available}, and licensed for non-commercial research purposes as specified in the Data Usage Agreement.

\section{Training details}
\textbf{Model architecture and training} We use a setup similar to~\citet{pmlr-v119-koh20a}. We use an Imagenet pre-trained Inception V3 backbone followed by a linear layer with 112 or 13 binary prediction heads (for \cub~and \chexpert~respectively) for the \ftoc~concept model. We use a linear \ctoy~label prediction model for \cub~and a 2-layer MLP with 128 hidden units for \chexpert. Similar to~\citet{pmlr-v119-koh20a}, for both datasets, we use the SGD optimizer with a fixed learning rate of 0.01 and a weight decay strength of 0.00004 for the independent \ftoc~and \ftoctoy~(joint and joint-sigmoid) models. For the independent \ctoy~model, we use a fixed learning rate of 0.001 with a weight decay strength of 0.00005.

\textbf{Intervention policies} We use the validation set to determine the optimal intervention order for \greedy~and for determining the optimal values of $\alpha$, $\beta$, and $\gamma$ for \ours. For the \cub~dataset, we use the train-val split as provided by~\citet{pmlr-v119-koh20a} and the standard test split. Followed by policy tuning, we retrain our models with the combined train+val set and report intervention results for the tuned policies on the test set. For \chexpert, we use custom train/val/test splits as described above. Unlike \cub, we don't retrain our models on the combined train+val set for \chexpert~as the train split is sufficiently big with 178731 examples.

\section{Additional Results}

% \subsection{Results on \chexpert\ data}

\subsection{Data efficiency of \ours}
\begin{figure}[!th]
    \centering
    \includegraphics[width=\linewidth]{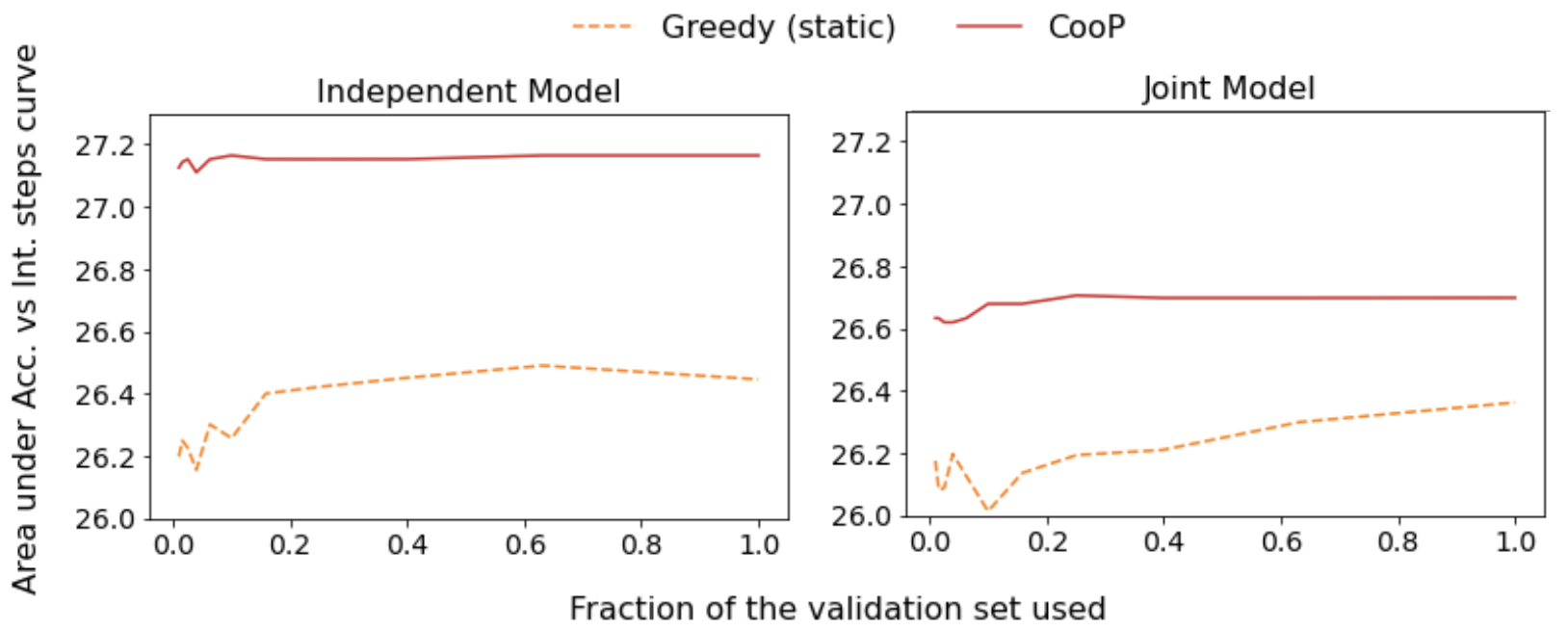}
    \caption{Data efficiency of \ours~compared with \greedy~evaluated by varying the size of validation set used for tuning the policies. y-axis denotes the Area under Top-1 Test Accuracy vs. Intervention Steps curve}
    \label{fig:data_eff}
\end{figure}

In Fig.~\ref{fig:data_eff}, we demonstrate a significant improvement in data efficiency obtained with \ours~as compared to \greedy. \ours~can consistently reach best-case performance with less than 20\% of validation data. Results are on the \cub~dataset.

\subsection{Ablation study}
\label{sec:ablation}
% \todo{update ablation to cost-sensitive data, each feature separately and all-together. leave greedy in, drop random and skyline}

\begin{figure}[!th]
    \centering
    \includegraphics[width=0.8\linewidth]{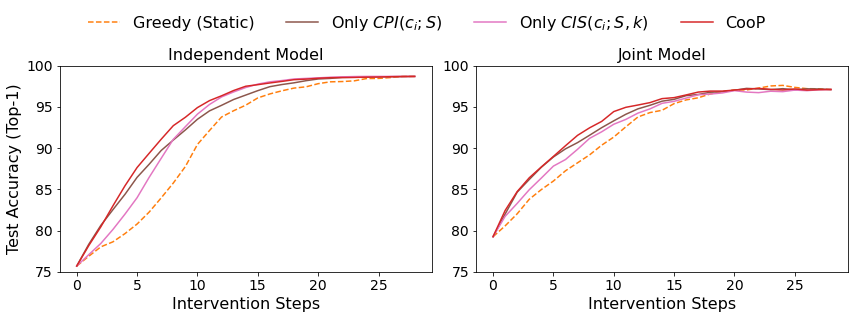}
    \includegraphics[width=0.8\linewidth]{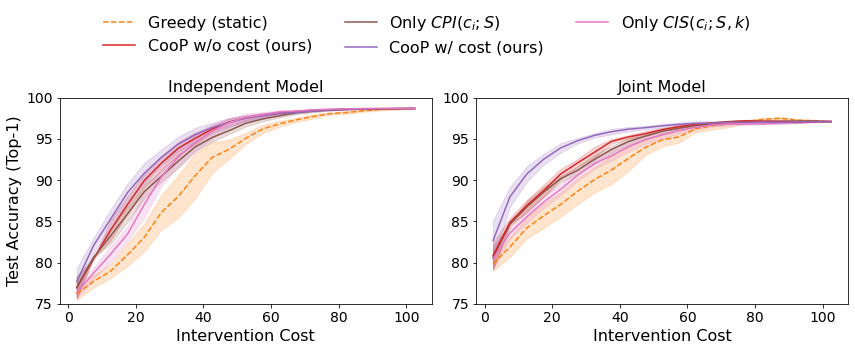}
    \caption{Ablation study on \cub}
    \label{fig:ablation}
\end{figure}

We performed additional experiments to gain insight into the manner in which policies such as \ours\ help improve outcomes in an intervention-efficient manner. Figure~\ref{fig:ablation} shows experiments on the \cub\ dataset using only one of the two factors: concept entropy, and change in label probability, rather than a combination of the 2 factors. We see that each factor allows us to learn instance-specific policies that are better than \greedy, but the combination beats both individual factors.

\section{Compute environment}
All experiments were performed in the cloud with individual jobs equipped
with an NVIDIA Tesla V100 and 32GB of memory.

\end{document}